\documentclass[sigconf]{acmart}
\AtBeginDocument{%
  \providecommand\BibTeX{{%
    \normalfont B\kern-0.5em{\scshape i\kern-0.25em b}\kern-0.8em\TeX}}}

\setcopyright{acmlicensed}
\acmConference[REML '23]{the 2023 ACM SIGIR Workshop on Retrieval-Enhanced Machine Learning}{July 27, 2023}{Taipei, Taiwan}
\acmYear{2023}
\copyrightyear{2023}
\acmDOI{}
\acmISBN{}
\acmPrice{}

\begin{document}

\title{Citations as Queries: \\ Source Attribution Using Language Models as Rerankers}

\author{Ryan Muther}
\email{muther.r@northeastern.edu}
\author{David A. Smith}
\email{dasmith@ccs.neu.edu}
\affiliation{%
  \institution{Northeastern University}
  \city{Boston}
  \state{MA}
  \country{USA}
}

\renewcommand{\shortauthors}{Muther and Smith}

\begin{abstract}
This paper explores new methods for locating the sources used to write a text, by fine-tuning a variety of language models to rerank candidate sources. After retrieving candidates sources using a baseline BM25 retrieval model, a variety of reranking methods are tested to see how effective they are at the task of source attribution. We conduct experiments on two datasets---English Wikipedia and medieval Arabic historical writing---and employ a variety of retrieval- and generation-based reranking models. In particular, we seek to understand how the degree of supervision required affects the performance of various reranking models. We find that semi-supervised methods can be nearly as effective as fully supervised methods while avoiding potentially costly span-level annotation of the target and source documents.
\end{abstract}



\received{25 April 2023}
\received[accepted]{14 June 2023}

\maketitle

\section{Introduction}

When reading a text, it is often useful to know which sources were used to write it. Knowledge of the sources used to write a news article, for example, can inform a reader of bias in how information in the article is reported. In historical domains, the sources used to write a document can both provide insight into how the author worked and what materials they had access to. We define the problem of determining the sources used to write a piece of text as that of source attribution. 

Researchers in natural language processing most often study source attribution in scientific papers, inferring links to referenced articles based on citations. Part of why this can be done so well is that modern citations follow a standardized format---often generated by required typesetting packages---that can be parsed by regular expressions or other simple methods. This comparative ease of data creation in turn allows the creation of large data sets for training fully supervised models for source attribution using the bibliographic information recovered from the citations. These forms of models tend to work best when there is a 1:1 correspondence between first printings of a work and papers. In more ambiguous domains, where potential sources can be redundant, these bibliographic models can often fail to disambiguate which of multiple possible editions of a work is the correct source. 

In order to properly retrieve sources in settings where the citations are harder to locate and potentially more ambiguous, creating fully annotated data in large amounts can sometimes be time consuming and require significant domain expertise. To circumvent this, we experiment with different levels of supervision in the models we use to retrieve potential sources. As we will see, semi-supervised methods can perform comparably to more annotation-intensive fully supervised models.

There are many forms of information one can use to aid in source attribution. Looking only at information present in the text itself, there are two principle forms of information about the sources: \textbf{text reuse} and \textbf{citation}. Text reuse is when an author directly copies material from their source, possibly involving more complex editing like summarization or paraphrase. This is common practice in highly intertexual domains like historical Arabic writing, where authors would often reuse each other's work. Citation involves the author explicitly telling the reader which source is being used, as one often sees in modern scientific writing or Wikipedia entries. Citations can have varying degrees of specificity, ranging from simply the author(s) and year, as one sees in some fields of scientific literature, to a more full-fledged citation including a title and page number, as is more common in many fields in the humanities. In some cases, the citation may take the form of a unique identifier, such as a (relative) URL or Wikipedia headword. 

Each of these forms of citation and reuse can be viewed as part of a broader spectrum of the form of relationship between a text and its sources. At one extreme, we have the case of Wikipedia, where the simplest form of citations to other Wikipedia articles often take the form of a link to the cited article by incorporating the name that article into the text of the citing article. At the opposite extreme, there is the reuse-heavy, often citationless, classical Arabic domain, where source attribution is more easily performed by recognizing the source text rather than any attribution on the part of the author. 

Different architectures can be used to model the process of source attribution from either the perspective of the author or that of the reader. From the perspective of the author, when deciding to write about a topic, one could imagine a process in which they select a source (retrieval) and then use the text of that source as the basis for their own writing (generation conditioned on that source). This process is similar to the process used by recently-proposed retrieval-augmented generation models in the work by Lewis et al. \cite{lewisRAG} From the perspective of the reader, the source retrieval problem is more like one of retrieval alone, as the reader doesn't need to create the target themselves, but can use it in the construction of a query to find sources.

We operationalize the problem of source retrieval by turning it into a two-stage retrieval and reranking problem. We first use a baseline retrieval model to retrieve candidate sources for a given target document. A second model of varying form is then used to rerank the possible sources retrieved by the baseline model.

This paper will be organized as follows; Section 2 will cover related work. Section 3 will provide an overview of the tasks and datasets we experiment with. Section 4 will cover the forms of models we use. Section 5 will go over the experiments and results. Section 6 will provide a discussion of the results as well as potential avenues for future work.

\section{Related Work}

Our work is closely related to work by others on combined retrieval and generation methods for question answering, citation suggestion, and literary evidence retrieval. For our experiments, we draw on a retrieval-augmented generation architecture first proposed by Lewis et al. \cite{lewisRAG} In contrast to Lewis et al, however, we are more focused on improving the model's retrieval performance rather than the generation performance, which is more pertinent when working on question answering as they are. Also of interest here is Mao et al's work on generation-augmented retrieval. \cite{mao2021generationaugmented} Their work focuses more on generating better queries by applying generative models than on using the generative model as a reranker after retrieving an initial candidate set. 

The problem of source attribution is quite similar to that of citation recommendation, which is usually set of as a tool for writers to find relevant citations to include. In \cite{citeRec}, for instance, Zhang and Zhu evaluate various forms of citation prediction model based on the similarity between the citation context and the citing paper on the task of predicting citations in PLOS ONE. While this sort of work is valuable for helping authors of scientific publications, it is limited in scope due to the way in which scientific papers tend to engage with their sources in a limited manner at the coarse level of the entire paper. In other more humanistic domains, the author often engages with the text of their sources directly and may reference multiple parts of a source, making our source attribution problem a more granular version of the commonly studied citation recommendation problem. 

Likely the most similar problem studied elsewhere is that of literary evidence retrieval as proposed by Thai et al in \cite{thai2022relic}. In RELIC, the goal is to retrieve the correct quoted passage from a known text based on its context in a work of literary analysis. The objective here is similar, but the relationship between the citing and cited texts may be more complex than direct quotation and the source text is not necessarily known, making a large space of possible sources than all passages of the source text.

\section{Datasets and Tasks}

\begin{table}[]
\begin{tabular}{l|l|l|}
Dataset    & Train & Test  \\ \hline
Wiki    & 116,038  & 12,880 \\ \hline
Maqrizi    & 162  & 19 \\ \hline
\end{tabular}
\caption{Sizes of datasets used in our experiments}
\label{tab:dataSizes}
\end{table}

We work with two datasets in this paper; Wikipedia links and two classical Arabic texts taken from the OpenITI corpus of digitized classical Arabic texts \cite{maxim_romanov_2019_3082464}. These two datasets in particular were chosen as the represent extremes in the spectrum of relationships between texts and their sources. In the case of the Wikipedia link task, the use of the source within a text requires very little modification of the source and could indeed be reduced to thinking of it as simply copying the headword of the source article. With Maqrizi, in contrast, the relationship between the text and its sources are much more complicated, with the source text often uncited or cited in ways that are difficult to recognize automatically, lacking any sort of standardized for like that found in other written traditions. Additionally the source may often be heavily edited or paraphrased in the target, further complicating the source-target relationship. Breakdowns of the sizes of training and test sets for the datasets can be seen in \ref{tab:dataSizes}. The texts used in the classical Arabic experiments are Maqrizi (d. 1442CE/845AH)'s \textit{Mawa'iz} and one of Maqrizi's sources, Ibn Abd al'Hakam (d. 871CE/257AH)'s \textit{Futuh Misr} and have been annotated by a domain expert on Maqrizi to create a dataset of 181 regions of shared text between the two works, some with and some without direct attribution in the form of citations by Maqrizi. When the other text is directly cited, the citation is separately marked in the annotations. The annotations are created based on the output of the text reuse detection algorithm Passim \cite{passim}, which operates by aligning sections of texts with a high number of shared character n-grams to find regions of shared material, which the domain expert refined to create the dataset of source-target pairs that we use in our experiments. Since the works in the corpus are so long, rather than aligning full texts, we cut the works up into 300-token chunks and align those. As a result, the alignments created by our annotator are at the chunk level, where chunk X of \textit{Futuh Misr} is a source for chunk Y of \textit{Mawa'iz}. The goal of the experiments with this dataset is to retrieve the proper source chunk for a given target chunk.

To create the dataset of Wikipedia citations, we collected a set of 150,000 links from Wikipedia articles to other Wiki articles where the link's anchor text was the name of the cited page from the citation dataset originally collected by Singh et al.\cite{colavizzaWiki} To better handle long articles, rather than treating individual articles as sources, retrieval and reranking is done at the section level, and any retrieved section from the correct source page is counted as relevant for evaluation purposes. For this dataset, the goal is to be able to retrieve a section from the cited page using the sentence with the link from the citing page, possibly with additional context from the surrounding content of the page. When we do include additional context, we add some number of sentences centered on the citing sentence, omitting any from a different, likely less relevant, section of the citing page. A harder version of this dataset could be created by requiring the the model to retrieve the exact cited section, but would limit available as not all Wikipedia links cite specific sections, with most citing a page as a single unit.

While one could accomplish the source attribution task in this simplified Wikipedia setting by using a simple lookup table of headwords from other Wiki pages, this task should not be discounted as uninteresting as a form of citation. The goal with the experiments on this data is to demonstrate that, as a form of citation, the kinds of models used in contexts where the relationship between the source and the target is more complicated than simple copying are also usable for this simplified domain.

\section{Models}

For our experiments we compare several kinds of models: a baseline retrieval model, as well as several forms of reranking models applied to the results of the baseline retrieval model; embedding similarity, generator-only, and generation-guided dense retrieval. We use each form of model to rerank candidate sources from among those retrieved by a baseline BM25 retrieval model. Each of these models is meant to test the usefulness of different forms of information in solving the source attribution task. The embedding similarity model is a baseline for how well untuned embedding models can solve this task. The generative models are used to examine how effectively generative models can learn to copy material from the source to the target as well as how the text is transformed in moving from the source to the target. Finally, the generation-guided retrieval models are meant to give insight into how well a task-specific tuned embedding model can perform relative to the untuned baseline. In contrast to \cite{lewisRAG}, where the retrieval-augmented generation architecture we are using was originally proposed, the goal is not to improve generation performance via retrieval, but rather to tune the embedding retrieval portion of the model by backpropagating the error from a combined retrieval and generation architecture. This, in theory, will minimize the distance in embedding space from target documents and those documents most useful for generating them, ideally the sources.

\subsection{Baseline Retrieval Model}

As noted above, the starting point of all of our experiments is a BM25 retrieval model used to retrieve possible sources, for which we use pyserini's implementation.\cite{pyserini} To allow the retrieval model to leverage information present in citations, we augment the source documents with bibliographic data, which is often otherwise not present in the source documents. In the case of Wikipedia, this takes the form of the article title. For the Maqrizi dataset, we augment sources with the author's name and the title of the source text, some combination of which is frequently be used by Maqrizi to indicate his sources when citations are present. 

\subsection{Embedding Similarity}

The simplest form reranks documents by descending order of cosine similarity of the representations of the source and target under a BERT embedding model trained on English and Arabic by Lan et al. \cite{lan_empirical_2020} As one might infer, the similarity is calculated as in Equation \ref{cosSim} where $t_{BERT}$ and $s_{BERT}$ are the BERT embeddings of the target and source respectively.

\begin{equation}\label{cosSim}
    sim = Cosine Similarity(t_{BERT},s_{BERT})
\end{equation}

\noindent This places source documents with more similar embeddings to the target document higher in the ranked list. The intuition behind this approach is that the sources used to write a text will be topically similar, meaning that the source documents may be nearby in embedding space. In practice, however, this intuition may not lead to better source retrieval performance in all domains, as experiments in Section 5 will show.

\subsection{Generator-Only}

The generator-only models rerank sources using the likelihood from a BART-based generation model \cite{BART} of some portion of the target (citing) document conditioned on the source (cited) document and the unmasked sections of the target. Unlike the embedding similarity method described in the previous section that only requires supervision at the document level, this method also requires the span of interest in the target to be annotated. The span of interest is the section of the target that we are interested in attributing to a particular source. At training time, this section of the target is replaces with <MASK> tokens and the BART model is trained to predict the masked span conditioned on concatenation of the masked target and source by minimizing the log likelihood of the masked span as in Equation \ref{BARTloss}.

\begin{equation}\label{BARTloss}
    L = -logp(t_{mask}|t_{obs},s)
\end{equation}

\noindent where $t_{mask}$ is the masked span in the target, $t_{obs}$ is the observed portion of the target, and $s$ is the source document. At inference time, the retrieved sources are reranked using the same loss, moving source documents that are more useful for generating the target text up in the ranked list of sources. For the Maqrizi dataset, since some annotated spans are often quite long, with the longest being 300 words (an entire input document), we truncate the masked sections to be at most 100 word pieces in length, leaving some of the target document to condition the generator.

Additionally, to test the feasibility of less annotation-heavy semi-supervised models, we also experiment with a semi-supervised version of BART where rather than using the known-correct source $s$ we substitute the top-ranking retrieved source from the baseline retrieval model $s'$, giving us 

\begin{equation}\label{BARTSemisupervisedloss}
    L = -logp(t_{mask}|t_{obs},s')
\end{equation}

\noindent as a loss function. In theory, a sufficiently strong baseline retrieval model will give BART enough correct documents to learn from, while learning to ignore erroneously retrieved irrelevant sources. This form of model, of course, still requires annotation at the span level in the target, but frees the annotator from locating the correct source for the target span of interest, which is often the more time-intensive portion of the task.

We can, of course, also relax this assumption that the human annotator needs to mark the span of interest as well by selecting a span of interest using some automatic method. In the case of Wikipedia, this is done by construction as the source links are themselves chosen automatically when constructing the dataset. For Maqrizi, such a dataset can be constructed using the raw passim alignments that the annotator used as the basis to construct the dataset in the first place. Rather than using the human-annotated target spans and verified source documents, the model can be trained on model-retrieved target spans and source documents.

\subsection{Generation-Guided Retrieval}

For the retrieval-augmented models, we base our approach on that of \cite{lewisRAG}, with some modifications for feasibility on the hardware we have access to in order to get around memory limitations of our hardware. To decrease the memory use of the large RAG models on the GPU, we reformulate the model so that rather than having to retrieve all the sources at once and run the generator with a batch size equal to the number of retrieved sources for a single target, each forward pass computes the loss with respect to a single retrieved source. We then accumulate gradients across all sources before updating the parameters with respect to all sources at once in the backward pass as described by Lewis et al in the RAG paper. We adopt the RAG-Sequence form of model proposed by Lewis et al. The loss for the model can be found in Equation \ref{ragLoss}, where $t$ and $s$ are the target and source documents, $q$ is the query used to retrieve sources for a given target, $t_i$ is the $i$th masked token in $t$. $p_{retriever}$ is the likelihood of retrieving source $s$ given query $q$, and $p_{gen}$ is like likelihood of generating the masked section of $t$ conditioned on the $q$ and $s$.

\begin{equation}\label{ragLoss}
    p_{RAG-Sequence}(t|q) = \sum_{s}p_{retriever}(s|q) \prod_{i}^{j}p_{gen}(t_i|t_{1:i-1},q,s)
\end{equation}

To further reduce the memory footprint, we switch to a smaller t5-small generator model (see \cite{t5}) which we freeze, so that the only portion of the model that is updated is the target-encoding BERT model used for the dense retrieval portion of RAG, rather than also updating the generator as in Lewis et al. This BERT model, guided by the retrieval-augmented generation training, is used to rerank exactly as in the untuned BERT case above. 

Similar to the BART model described in the previous section, the RAG training also involves annotating the target document to define the span of interest, which allows the generative portion of RAG to be used to encourage the retrieval model to move the representations of target documents closer to their sources. \footnote{We also experiment with a semi-supervised BERT-based reranker trained on pairs of documents and a single query to predict the relevance scores for the documents, and optimize a hinge loss based on the predicted scores as in \cite{DBLP:journals/corr/DehghaniZSKC17}, but performance was lacking so we do not report those results here.}

\section{Experiments}

We divide our experiments into two main sections; retrieval-oriented and generation-oriented. The retrieval experiments are meant to explore how effective different forms of models with varying degrees of supervision are at solving the problem of source attribution, and are evaluated using Recall@10 and Mean Reciprocal Rank. The generation experiments, in contrast, are intended to try to understand the degree to which generative models are capable of learning to copy from source documents, rather than relying on the ability to fill in masked text using only the surrounding context using the understanding of language gained during pretraining. To this end, we evaluate the accuracy of the generator when tasked with predicting the masked text, rather than employing the generator as a reranker as in the retrieval experiments. This is of particular interest in the domain of Wikipedia, as it was part of the model's pretraining dataset, so data leakage from the pretraining data may have occurred. As will be deomstrated, this is not in fact the case.

\subsection{Retrieval Experiments}

\begin{table}[]
\begin{tabular}{l|l|l|l|l|}
Dataset & Model    & Supervision & R@10 & MRR  \\ \hline
Wiki    & Baseline & N/A & .64  & .478 \\ \hline
Wiki    & +BERT     & N/A & .14  & .068 \\ \hline
Wiki    & +BART     & Pair- and Span-level & .97  & .927 \\ \hline
Wiki & +BART-Semi     & Span-Level Only & .94  & .895 \\ \hline
Wiki    & +RAG      & Unsupervised & .15  & .077 \\ \hline
Maqrizi & Baseline & N/A & .84  & .680 \\ \hline
Maqrizi & +BERT     & N/A & .36  & .300 \\ \hline
Maqrizi & +BART     & Pair- and Span-level & .95  & .948 \\ \hline
Maqrizi & +BART-Semi     & Span-Level Only & .95  & .947 \\ \hline
Maqrizi & +BART-Passim     & Semi-supervised & .89  & .897 \\ \hline
Maqrizi & +RAG      & Unsupervised & .36  & .300 \\ \hline
\end{tabular}
\caption{Results for Reranking Experiments with various models and datasets}
\label{tab:rerankScore}
\end{table}

We will now describe our experiments so far on the Wikipedia and Maqrizi datasets. As a baseline, we use a simple bag-of-words BM25 retrieval model as implemented in pyserini.\cite{pyserini} For Wikipedia, the baseline retrieval model attains a recall at 10 of .64 and MRR of .478. If we then use these retrieval results as input to a BART-based (i.e. generation only) reranking model, which has been trained to generate the link text conditioned on the masked target text and complete source text, the recall at 10 increases to .97 and the MRR to .927. Both the unsupervised and supervised retrieval-based models, untrained BERT and RAG, display vastly worse performance than the baseline model, with RAG very slightly outperforming BERT. It may be that the compromises made to make training and inference with RAG feasible also made RAG unable to actually learn how to solve the problem. It seems likely that being able to tune the generator during RAG's fine tuning is important for learning to copy from the retrieved source, and that simply reweighting the retrieval scores based on the frozen generator's likelihood is insufficient for properly reranking sources. It may also be that the poor performance of BERT acts as a limiting factor on RAG's performance, judging by the very similar performance between BERT and RAG. Training with partial supervision performs almost as well as a fully supervised model with .94 recall at 10 and .895 MRR, despite the worse performance of the baseline retrieval model. 

On the Maqrizi dataset, the baseline retrieval model attains a recall at 10 of .84 and an MRR of .680. Reranking with BERT without any further pretraining for the task actually degrades performance, decreasing recall at 10 to .36 and MRR to .30. As with the Wikipedia data, reranking with a RAG model also produces similar results, again likely due to the lack of fine-tuning for the generator preventing the model from meaningfully learning how to rerank sources. Again, similar to Wikipedia, reranking with a purely generative approach significantly improves performance, reaching .95 recall at 10 and an MRR of .948. Interestingly, as with the Wikipedia dataset, most of this performance increase is maintained if we switch the training data from fully supervised training to a semi-supervised setup with only span-level supervision in the target documents, with a very minor decrease in MRR to .947. This continues to hold true even when one relaxes the constraint that the spans to generate also be human-annotated, as one can see from the BART-Passim model, where the model outperforms the retrieval baseline both in terms of MRR and Recall@10. Due to the extreme length of the masked sections in this data set, a similar evaluation for the predictive accuracy of BART without conditioning on the source document would likely be uninformative as the odds of correctly predicting an entire span of 100 subword tokens exactly correctly would be much lower, making the exact match evaluation as performed on the Wikipedia data in the next section much less informative.

\subsection{Generation Experiments}

\begin{table}[]
\begin{tabular}{l|l|}
Training Data    & Generation Accuracy \\ \hline
None & 0 \\ \hline
Target-Only     & .170 \\ \hline
Target and Source     & .714 \\ \hline
Target and Semi-Supervised Source     & .709 \\ \hline
\end{tabular}
\caption{Results for Reranking Experiments with various models and datasets}
\label{tab:generationScore}
\end{table}

It may be that the improved performance comes from the generative ability imparted by BART's pretraining rather than the learned ability of the fine-tuned model to copy from sources. Furthermore, for the Wikipedia experiments, the training and test sets are themselves a part of what BART was originally pretrained on, making data leakage between the pretraining dataset and this downstream task possible. To test that this isn't artificially inflating the models source reranking performance, we measure the predictive accuracy of three forms of BART-only models at the task of filling in the masked link text on the Wikipedia dataset. The results for these experiments can be seen in Table \ref{tab:generationScore}. First, we test completely untuned BART-base and test using the text of the target only. Secondly, we use the same test input, but fine tune the model on the link prediction task. Finally, we do the full conditioning on the masked target and source section as described above. The untuned model was completely unable to predict the link text, attaining 0\% accuracy, while the target-only model achieved .170 accuracy. The combined source-target model, which is the fully supervised setup from the previous section, achieved .714 accuracy. As this shows, this training process is, as we hope, teaching the model to copy from the source document, rather the ability to rerank being a side effect of the presence of Wikipedia in BART's pretraining data. This predictive performance is largely maintained when we use the highest-ranked BM25 retrieved source in place of the known-correct source, with a very minor decrease to .709 accuracy.

\section{Discussion and Future Work}

The experiments discussed in the previous section provide insight into how well different forms of large language model can be used to solve the problem of source attribution, as well as the importance of task-specific fine tuning. While it is clear that generative models like BART are capable of learning more complex source-target relationships than simple copying as evidenced by the strong performance of BART on both data sets. However, the required annotation to train such a model makes it unattractive as a general solution to the problem, though they may be useful for small specialized domains like that of Maqrizi in this case. In contrast, the poor performance of both the untuned BERT and partially-frozen RAG models indicates that, clearly, these models perform poorly without additional training, as one would expect. RAG in particular suffers despite the additional complexity of the generative portion due to its lack of fine-tuning to learn how to constructed the masked text from the source document. Semi-supervised methods, however, may be an attractive way to make progress on improving source ranking without time-consuming annotation.

There are several interesting avenues for potential future research on this topic. Further work on unsupervised methods like RAG may be appealing with access to better hardware capable of running a more complete version of the model. As the semi-supervised results show, it is also potentially fruitful to apply semi-supervised methods to this problem to avoid costly annotation for, as it seems, a small loss in performance. It would be worth evaluating this approach on a larger dataset to see if the conclusions we draw from the Maqrizi test set generalize beyond that small setting. Indeed, one benefit of moving to a semi-supervised approach where both the spans and document pairs don't require human annotation is that all the human annotated data can be used for evaluation. We avoid doing so here in the interest of fair comparison between the various forms of model that do require human annotated training data. It would also be worthwhile to examine the performance of finetuned BERT trained to embed sources and targets more closely as a potential reranker.

Additionally, as Maqrizi and Wikipedia represent very extreme notions of what source-target relationships look like, it would be valuable to find another dataset where citation is more formally structured than Maqrizi, but less than in Wikipedia as a third use case. For instance, once could imagine looking the the work of 19th century philosopher J. S. Mill and his sources, which not only have a citations more along the lines of what one sees in modern writing, but would also allow one to examine the utility of these methods in a crosslingual setting, as he often cites sources in languages other than English that are also digitized. One could also attempt to look at Wikipedia's citations to other sources, like the Internet Archive or Google Books.

\begin{acks}
The authors would like to thank Mathew Barber at the Aga Khan University's Centre for Digital Humanities for annotating the Maqrizi dataset, as well as the anonymous reviewers for their helpful feedback. This paper is part of a project that has received funding from the European Research Council (ERC) under the European Union's Horizon 2020 research and innovation program (Grant agreement No. 772989)
\end{acks}

\bibliographystyle{ACM-Reference-Format}
\bibliography{custom}

\end{document}